\documentclass[11pt,a4paper]{article}
\usepackage[hyperref]{acl2020}
\usepackage{times}
\usepackage{latexsym}
\usepackage{graphicx}
\usepackage{subfig}
\usepackage{url,comment}
\usepackage{multirow}
\usepackage{xspace}
\usepackage{booktabs}

\aclfinalcopy % Uncomment this line for the final submission
\def\aclpaperid{2211} %  Enter the acl Paper ID here

\newcommand{\presec}{\vspace{0.0in}}
\newcommand{\postsec}{\vspace{0.0in}}
\newcommand{\presub}{\vspace{0.0in}}
\newcommand{\postsub}{\vspace{0.0in}}

\newcommand{\postsubsub}{\vspace{0.0in}}

\usepackage{makecell}

\newcommand{\amt}{\textsc{EBG}\xspace}
\newcommand{\blogs}{\textsc{BLOG}\xspace}

\newcommand{\svm}{\textsc{SVM}\xspace}
\newcommand{\rfc}{\textsc{RFC}\xspace}
\newcommand{\knn}{\textsc{KNN}\xspace}
\newcommand{\nueralnetworks}{\textsc{ANN}\xspace}
\newcommand{\naivebayes}{\textsc{GNB}\xspace}

\newcommand{\gptsmall}{\textsc{GPT-2 117M}\xspace}
\newcommand{\gptlarge}{\textsc{GPT-2 345M}\xspace}
\newcommand{\bertsmall}{\textsc{BERT base}\xspace}
\newcommand{\bertlarge}{\textsc{BERT large}\xspace}

\newcommand{\dspan}{\textsc{DS-PAN17}\xspace}
\newcommand{\snpan}{\textsc{SN-PAN16}\xspace}
\newcommand{\mutantx}{\textsc{Mutant-X}\xspace}

\title{A Girl Has A Name: Detecting Authorship Obfuscation}

\author{Asad Mahmood  \hspace{10mm} Zubair Shafiq  \hspace{10mm}  Padmini Srinivasan\\\\
The University of Iowa \\
  {\small \tt \{asad-mahmood,zubair-shafiq,padmini-srinivasan\}@uiowa.edu} }

\date{}

\begin{document}
\maketitle
\begin{abstract}
  \textit{Authorship attribution} aims to identify the author of a text based on the stylometric analysis.
\textit{Authorship obfuscation}, on the other hand, aims to protect against authorship attribution by modifying a text's style.
In this paper, we evaluate the stealthiness of state-of-the-art authorship obfuscation methods under an adversarial threat model.
An obfuscator is \textit{stealthy} to the extent an adversary finds it challenging to detect whether or not a text modified by the obfuscator is obfuscated -- a decision that is key to the adversary interested in authorship attribution.
We show that the existing authorship obfuscation methods are not stealthy as their obfuscated texts can be identified with an average F1 score of 0.87.
The reason for the lack of stealthiness is that these obfuscators degrade text smoothness, as ascertained by neural language models, in a detectable manner. 
Our results highlight the need to develop stealthy authorship obfuscation methods that can better protect the identity of an author seeking anonymity.

\end{abstract}

\section{Introduction}

%% - Authorship Attribution
Authorship attribution aims to identify the author of a text using stylometric techniques designed to capitalize on differences in the writing style of different authors. 
Owing to recent advances in machine learning, authorship attribution methods can now identify authors with impressive accuracy \cite{abbasi2008writeprints} even in challenging settings such as cross-domain \cite{overdorf2016blogs} and at a large-scale  \cite{narayanan2012feasibility,ruder2016character}.
Such powerful authorship attribution methods pose a  threat to privacy-conscious users such as journalists and activists who may wish to publish anonymously \cite{times2018part, amazon18employeeletter}.

Authorship obfuscation, a protective countermeasure, aims to evade authorship attribution by obfuscating the writing style in a text.
Since it is challenging to accomplish this manually, researchers have developed automated authorship obfuscation methods that can evade  attribution while preserving semantics \cite{PAN@CLEF}.
However, a key limitation of prior work is that authorship obfuscation methods do not consider the adversarial threat model where the adversary is ``obfuscation aware'' \cite{karadzhov2017case,potthast2018overview,mahmood2019girl}. 
Thus, in addition to evading attribution and preserving semantics, it is important that authorship obfuscation methods are ``stealthy'' -- i.e., they need to hide the fact that text was obfuscated from the adversary.

In this paper, we investigate the stealthiness of state-of-the-art authorship obfuscation methods.
Our intuition is that the application of authorship obfuscation results in subtle differences in text smoothness (as compared to human writing) that can be exploited for obfuscation detection. 
To capitalize on this intuition, we use off-the-shelf pre-trained neural language models such as BERT and GPT-2 to extract text smoothness features in terms of word likelihood.
We then use these as features to train supervised machine learning classifiers. 
The results show that we can accurately detect whether or not a text is obfuscated.

Our findings highlight that existing authorship obfuscation methods themselves leave behind stylistic signatures that can be detected using neural language models.
Our results motivate future research on developing stealthy authorship obfuscation methods for the adversarial threat model where the adversary is obfuscation aware.

\vspace{.1in}
Our key contributions are as follows:

\begin{itemize}
  \item We study the problem of obfuscation detection for state-of-the-art authorship obfuscation methods. This and the underlying property of stealthiness has been given scant attention in the literature.
  We also note that this problem is potentially more challenging than the related one of synthetic text detection since most of the original text can be retained during obfuscation. 
  
  \item We explore 160 distinct BERT and GPT-2 based neural language model architectures designed to leverage text smoothness for obfuscation detection. 
  
  \item We conduct a comprehensive evaluation of these architectures on 2 different datasets. 
  Our best architecture achieves F1 of 0.87, on average, demonstrating the serious lack of stealthiness of existing authorship obfuscation methods.
\end{itemize}

\noindent \textbf{Paper Organization:} The rest of this paper proceeds as follows.
Section \ref{sec: related work} summarizes related work on authorship obfuscation and obfuscation detection.
Section \ref{sec: approach} presents our proposed approach for obfuscation detection using neural language models.
Section \ref{sec: setup} presents details of our experimental setup including the description of various authorship obfuscation and obfuscation detection methods.
We present the experimental results in Section \ref{sec: results} before concluding.  
The relevant source code and data are available at \url{https://github.com/asad1996172/Obfuscation-Detection}.

\section{Related Work}
\label{sec: related work}
In this section, we separately discuss prior work on authorship obfuscation and obfuscation detection.

\subsection{Authorship Obfuscation}

Given the privacy threat posed by powerful authorship attribution methods, researchers have started to explore text obfuscation as a countermeasure.
Early work by Brennan et al. \shortcite{brennan2012adversarial} instructed users to manually obfuscate text such as by imitating the writing style of someone else. 
Anonymouth \cite{mcdonald2012use,mcdonald2013anonymouth} was proposed to automatically identify the words and phrases that were most revealing of an author's identity so that these could be manually obfuscated by users.
Follow up research leveraged automated machine translation to suggest alternative sentences that can be further tweaked by users \cite{almishari2014cosn,keswani2016author}.
Unfortunately, these methods are not effective or scalable because it is challenging to manually obfuscate text even with some guidance.

Moving towards full automation, the digital text forensics community \cite{PAN2018overview} has developed rule-based authorship obfuscators 
\cite{mansoorizadeh2016author,karadzhov2017case, dcastrocastro2017case}.  
For example, Karadzhov et al. \shortcite{karadzhov2017case} presented a rule-based obfuscation approach to adapt the style of a  text towards the ``average style'' of the text corpus.
Castro et al. \shortcite{dcastrocastro2017case} presented another rule-based obfuscation approach to ``simplify'' the style of a  text.

Researchers have also proposed search and model based approaches for authorship obfuscation. 
For example, Mahmood et al.  \shortcite{mahmood2019girl} proposed a genetic algorithm approach to ``search'' for words that when changed, using a sentiment-preserving word embedding, would have the maximum adverse effect on authorship attribution.
Bevendorff et al. \shortcite{bevendorff2019heuristic} proposed a heuristic-based search algorithm to find words that when changed using operators such as synonyms or  hypernyms, increased the stylistic distance to the author's text corpus.
Shetty et al. \shortcite{shetty2018a4nt} used Generative Adversarial Networks (GANs) to ``transfer'' the style of an input text to a target style.
Emmery et al. \shortcite{emmery2018style} used auto-encoders with a gradient reversal layer to ``de-style'' an input text (aka style invariance).

\presub
\subsection{Obfuscation Detection}
\postsub
Prior work has successfully used stylometric analysis to detect \textit{manual} authorship obfuscation \cite{juola2012detecting,afroz2012detecting}. 
The intuition is that humans tend to follow a particular style as they try to obfuscate a text.
In a related area, Shahid et al. \shortcite{shahid2017accurate} used stylometric analysis to detect whether or not a document was ``spun'' by text spinners. 
We show later that these stylometric-methods do not accurately detect more advanced automated authorship obfuscation methods.

There is increasing interest in distinguishing synthetic text generated using deep learning based language models such as BERT and GPT-2 from human written text.
Using contextual word likelihoods, as estimated using a pre-trained language model  \cite{radford2019language}, Gehrmann et al. \shortcite{gehrmann2019gltr} were able to raise the accuracy of humans at detecting synthetic text from 54\% to 72\%.
Zellers et al. \shortcite{zellers2019defending}
% \cite{zellers2019defending} \citeauthor{zellers2019defending} \citeyear{zellers2019defending} 
%took this a step further and 
showed that a classifier based on a language model can accurately detect synthetic text generated by the same language model.
However, the detection accuracy degrades when different language models are used to generate and to detect. 
Bakhtin et al. \shortcite{bakhtin2019real} also showed that the detection accuracy degrades when the synthetic text is generated using a language model trained on a different corpus.

In summary, recent research has leveraged language models to detect their generated synthetic text. 
However, in obfuscation we start with human written text and make modifications such that text semantics is still preserved.
This is in part achieved by retaining chunks of the original writing.
Thus, the quirks of the obfuscator will be mingled in unpredictable proportions and ways with the author's original writing style.
This makes the detection of obfuscated text different and potentially more challenging than synthetic text detection.
To the best of our knowledge, this work presents the first systematic study of the detection of automatically obfuscated text.

\presec
\section{Proposed Approach}
\label{sec: approach}

\begin{figure*}[!t]
	\centering
	\includegraphics[width=\linewidth]{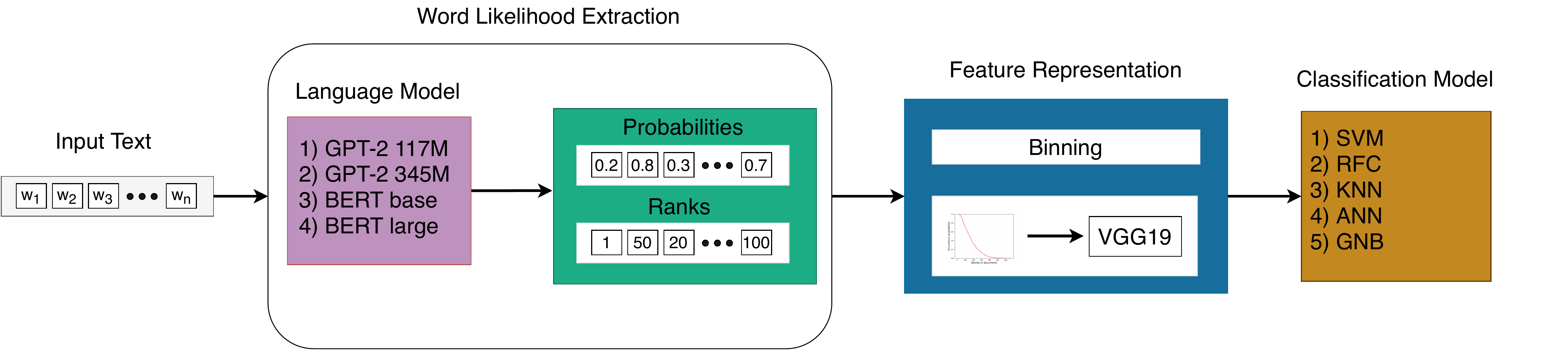}
	\caption{Pipeline for obfuscation detection}
    \label{pipeline}
\end{figure*}
% \presub

\subsection{Intuition}
\postsub
An automated authorship obfuscator changes the input text so that it evades authorship attribution while preserving semantics.
The quality and smoothness of automated text transformations using the state-of-the-art obfuscators differ from that of human written text \cite{mahmood2019girl}.
Therefore, the intuition behind our obfuscation detectors is to exploit the differences in text smoothness between human written and obfuscated texts. 
We capture text smoothness using powerful pre-trained context aware neural language models.\footnote{BERT: {https://ai.googleblog.com/2018/11/open-sourcing-bert-state-of-art-pre.html}; \\GPT-2: {https://openai.com/blog/better-language-models}}
A text with a relatively greater proportion of high likelihood words is likely to be more smooth.

% \presub
\subsection{Detector Architectures}
\postsub
Figure \ref{pipeline} shows the pipeline of our method for detecting whether or not a given text is obfuscated.
First, a language model is used to extract the likelihood (in the form of probability or rank) for each word in the text.
Second, these likelihoods are used to build a smoothness representation for the text.
This is input to a supervised machine learning model that is trained to classify the text as human written or obfuscated.
The three steps correspond to three significant architectural dimensions of our detectors with multiple algorithmic options in each dimension.
%Given a document, we first use language models to extract the likelihood of each word.
%
%Then, we use these likelihood values to extract features which capture the smoothness of text.
%
%In the end, we use these features to perform classification.
%
Combinations of choices along each dimension yield different architectures that can be used by an adversary to detect obfuscated documents.
We detail each dimension next.
% \presubsub

\subsubsection{Word likelihood extraction}
\postsubsub

Given a word sequence, language models are designed to predict the next word.
%
%Traditionally, given a sequence of words, language models are trained to predict the next word.
%
They do this by building contextual models of word occurrences as probability distributions over the full vocabulary.
Then some heuristic %based on these probability distributions, 
is used to pick the next word e.g., select the word with the highest probability.
In our case, instead of word prediction, we extract the likelihood from the language model (either as a probability or as a rank) for each word in the text given its context.
%
%For example, if `cat' occurring at a given position in the document has the second highest probability as per the model then it's rank is 2.
%
%So we use a language model to create probability distributions for every word and then extract 1) the occurrence probability of the actual word and 2) the rank of the actual word.

%
The language model has a critical role.
Thus, we use neural language models with deep architectures and trained on large amounts of data which are better at identifying both long-term and short-term context.
In order to imitate an adversary who may not have the significant resources needed to train such models, we use off-the-shelf pre-trained neural language models.
Specifically, we choose well-known context-aware neural language models GPT-2 \cite{radford2019language} and BERT \cite{devlin2018bert}.
We choose both as they use different approaches.
GPT-2 has been shown to perform better than BERT \cite{gehrmann2019gltr} at synthetic text detection, with word rank giving higher performance than word probability. 
Their relative merit for obfuscation detection is unknown.

\vspace{0.05in} \noindent \textbf{1) GPT-2.}
GPT-2 released by Open AI in 2019 uses at its core, a variation of the ``transformer'' architecture, an attention based model \cite{vaswani2017attention} and is trained on text from 45 million outbound links on Reddit (40 GB worth of text).
We use GPT-2 to compute the conditional probability for word \textit{i} as $p(w_i | w_{1...i-1})$.
The position of $w_i$ in the sorted list (descending order of probability) of vocabulary words gives the word rank.
The authors \cite{radford2019language} trained four versions of GPT-2 differing in architecture size.
Of these, we used the small and medium versions containing 117M and 345M parameters, respectively.
The authors eventually also released a large version containing 762M parameters and a very large version containing 1542M parameters.\footnote{https://openai.com/blog/gpt-2-6-month-follow-up/} 
We did not use them because only the small and medium versions were released at the time of our experimentation. 

\vspace{0.05in} \noindent \textbf{2) BERT.} BERT released by Google in 2018 is also based on ``Transformers''. 
It is trained on text from Wikipedia (2.5B words) and BookCorpus (800M words).
BERT considers a bi-directional context unlike the uni-directional context considered by GPT-2.
Thus, in BERT the conditional occurrence probability for word \textit{i} is $p(w_i | w_{i-k...i-1}, w_{i+1...i+k})$ where k is the window size on each direction.
Rank is computed in the similar way as GPT-2.
We use both pre-trained BERT: \bertsmall with 110M parameters and \bertlarge with 340M parameters.

We implement likelihood extraction for both GPT-2 and BERT, using code made available by the Giant Language Model Test Room (GLTR) tool.\footnote{https://github.com/HendrikStrobelt/detecting-fake-text}

%
% \presubsub
\subsubsection{Feature Representation}
\postsubsub
We experiment with two different representations of smoothness. %
Each is explored with occurrence probabilities and with ranks.

\vspace{0.05in} \noindent \textbf{1) Binning based features: }
Text smoothness is represented by the  likelihood of words in text.
A text with a greater proportion of high likelihood words is likely to be smoother.
%
% 
%We consider word likelihoods by aggregating the information into bins.
%as the feature representation would be effected by noise.
%
We aggregate this information using fixed size bins representing different likelihood ranges. 
%To reduce the effect of noise, we use fixed sized binning.
%
For probabilities we create bin sizes of 0.001, 0.005 and 0.010. 
For ranks we create bin sizes of
10, 50 and 100.
Thus for example, one feature representation is to consider bins of ranks from 0 to 10, 11 to 20, 21 to 30 etc.
Each bin contains the proportion of words in the document with likelihood in that range.
% For example, for bin size = 0.1, first value of the resulting feature vector would be the average number of words having probabilities between 0 and 0.1 and so on.
%

\vspace{0.05in} \noindent \textbf{2) Image based features: }
Since the word likelihood values received from language models are in essence signals, we explore signal detection approaches as well.
For example, for audio classification
\cite{hershey2017cnn} store plots of the log-mel spectogram of the audios as images and then apply image classification methods.
VGG \cite{simonyan2014very}, was one of the top performers of the different classifiers they tested.
Inspired by them, we explore obfuscation detection via image classification.
Specifically, we explore a transfer learning approach wherein we use the VGG-19 classifier\footnote{https://keras.io/applications/\#vgg19} trained for image classification on ImageNet dataset\footnote{ http://www.image-net.org/}.
For our method, we sort the extracted likelihood values for the text in descending order and then plot these values saving it as an image.
This image is then processed by the pre-trained VGG-19.
We extract the document's \footnote{Terms `text' and `document' are used interchangeably} representation from the last \textit{flatten} layer of VGG-19 (before the fully connected layers) as it contains high-level information regarding edges and patterns in the image.
We expect this resulting feature representation vector to capture information regarding text smoothness.
%
% \begin{figure}
%   \centering
%   \includegraphics[width=\linewidth]{images/sorted_orig_ex_prob.pdf}
%   \caption{Plot of sorted occurrence probabilities (using \bertsmall model) for a human generated document.}\label{sorted_probs_plot}
% \end{figure}
% This is done for every document.
% \presubsub
\subsubsection{Classification}
\postsubsub
We experiment with Support Vector Machine (\svm) with a linear kernel, Random Forest Classifier (\rfc) an ensemble learning method, K Nearest Neighbor (\knn) which is a non-parametric method, Artificial Neural Network (\nueralnetworks) which is a parametric method, and Gaussian Naive Bayes (\naivebayes) which is a probabilistic method.
All classifiers are trained using default parameters from scikit-learn\footnote{https://scikit-learn.org/stable/} except for \nueralnetworks, where we use \textit{lbfgs} solver instead of \textit{adam} because it is more performant and works well on smaller datasets.

% \presubsub
\subsubsection{Detection Architectures}\label{all_architectures}
\postsubsub
Options selected for each dimension combine to form a distinct obfuscation detection architecture.
With 4 language models giving probabilities or ranks as output, 4 features (3 binning based features and 1 image based feature) and 5 different classifiers we experiment with a total of 160 distinct architectures.
The assumption here is that a determined adversary will similarly look for the most effective obfuscation detector.

%
%An example of a specific architecture is: using \gptsmall language model with occurrence probabilities using image based features with an \svm classifier.

% \presec
\section{Experimental Setup}
\label{sec: setup}
\postsec
%We first present the obfuscation methods tested for stealthiness and then our data.
%
%Next we present the detection methods tested including ones that are external to our architecture.
%
%We first talk about the two datasets being used for experiments.
%
%Then we briefly discuss the authorship obfuscation approaches being used.
%
%After that we discuss the baselines and other methods that can be used for obfuscation detection.
%
%In the end, we discuss the evaluation metrics used to measure the performance.
% \padmini{See we are creating a dataset by passign the 2 datasets through different obfusctors. that part needs clear explanation.  so we start with blogs and amt and create these.  how we create these is partially identified. need a pic.}

% \presub
\subsection{Authorship Obfuscation Approaches}\label{obf_approaches}

As state-of-the-art automated authorship obfuscators we identified the top two systems \cite{potthast2018overview} from PAN, a shared CLEF task.\footnote{https://pan.webis.de/}
We also chose Mutant-X, a search based system presented in \cite{mahmood2019girl}, which shows better performance than the PAN obfuscation systems. These are detailed next.

\vspace{0.05in} \noindent \textbf{Document Simplification \cite{dcastrocastro2017case}}.
This approach obfuscates by applying rule-based text simplifications on the input document.
The process is as follows. %consists of three steps.
1) If the number of contractions in the document is greater than the number of expansions, then replace all contractions with expansions otherwise replace all expansions with contractions.
2) Simplify by removing parenthetical texts that do not contain any named entity, discourse markers or appositions.
3) Replace words with synonyms that haven't been already used in the text.
We implement this approach and refer to it as \dspan.

\vspace{0.05in} \noindent \textbf{Style Neutralization \cite{karadzhov2017case}}.
This system is also a rule-based text obfuscator.
First they calculate the average values  for the whole corpus for stylometric features such as stopword to non stopword ratio, punctuation to word count ratio and average number of words per sentence.
Next, they calculate the values of same stylomteric features for the input document.
Finally, using text transformation rules (e.g., replace ! with !!, merge or split sentences etc.) they move the document's stylometric feature values towards the corpus averages.
We evaluate this approach using the code provided by the authors and refer to it as \snpan.

\vspace{0.05in} \noindent \textbf{\mutantx \cite{mahmood2019girl}}
This system uses a genetic algorithm (GAs) in combination with an authorship attribution system to identify words that when changed would have the highest positive effect towards obfuscation.
Text transformations are done using a sentiment preserving variation of Word2Vec \cite{yu2017refining}. 
The authors present two versions:
\mutantx writeprintsRFC, built using a traditional machine learning based authorship attribution system and \mutantx embeddingCNN, built using a deep learning based authorship attribution system.
We evaluate \mutantx embeddingCNN using code made available by authors.
%\padmini{is the CNN version one of the best performers in our earlier study? if so say so}
%{\bf did you make this code public? \asad{No, should we say that we implemented it?}}

\subsection{Data}

We use the two data collections which were used by \cite{mahmood2019girl}.

\vspace{0.05in} \noindent 1) \textbf{Extended Brennan Greenstadt corpus}. This text corpus from \cite{brennan2012adversarial} contains 699
documents written by 45 unique authors.
Documents are mainly academic in nature but they do not contain any citations and section headings and have under 500 words, as instructed by data collectors.
% to write documents which
% do not contain any citations, section headings and
% have less than 500 words.
%
% On average, every author submitted 15 documents.
%
We refer to this collection as \amt.

\vspace{0.05in} \noindent 2) \textbf{Blog authorship corpus}. 
This text corpus which is from \cite{schler2006effects} contains more than 600,000 blogger.com blog posts written by 19,320 unique authors.
% This is a corpus containing more than 600,000 blog posts
% by 19,320 unique authors from blogger.com.
%
% Every author has 35 blog posts on average.
%
%
These texts are more casual in nature and are written without any instructions.
%
% Instead of considering the whole corpus, we filter it by extracting 100 authors who have exactly 100 blogs each. \padmini{why is this 100 authors point important?? \asad{No reason, was just stating what was done}}
%
We refer to this collection as \blogs.

The authors in \cite{mahmood2019girl} tested their GA obfuscators using 49 \amt and 200 \blogs documents written by 10 authors.
%

%%%%%%%%%%%%%%%%%%%%%%%%%%%%%%%%%%%%%%%%%%%%%%%%%%%%%%
% \presubsub
\subsubsection{Obfuscated versus Evaded Datasets}\label{obfandevaded}

%%%%%%%%%%%%%%%%%%%%%%%%%%%%%%%%%%%%%%%%%%%%%%%%%%%%%%
A document is considered obfuscated if it has been processed by an authorship obfuscation tool.
Thus we processed each document in the 49 \amt and 200 \blogs sets using each of the three obfuscation systems in turn (DS-PAN17, SN-PAN16 and MUTANT-X embeddingCNN). 
This resulted in 127 obfuscated \amt documents and 571
obfuscated \blogs documents.\footnote{DS-PAN17, SN-PAN16 and MUTANT-X embeddingCNN respectively created 49, 49 and 29 obfuscated documents for \amt whereas for \blogs they created 200, 200 and 171 obfuscated documents. Lower number of documents generated by MUTANT-X is due to the fact that it only obfuscates documents that are correctly classified by the attribution classifier it uses.}

An obfuscated document is viewed as an evaded document if it successfully evades authorship attribution.
Using the writeprintsRFC authorship attributor \cite{mahmood2019girl} for this purpose, we process all obfuscated documents (127 and 571 for \amt and \blogs respectively). 
This resulted in 59 evaded \amt documents (out of 127) and 184 evaded \blogs documents (out of 571). 

Despite the type and extent of transformations made,  the remaining obfuscated documents did not evade the attributor.
The evaded documents are those where the modification strategy somehow crossed an implicit threshold for evading authorship attribution.
%which does not happen for all obfuscation instances.
Thus, we surmise that the evaded documents are likely to be relatively less smooth. % also expect that evaded 
%Our intuition is that in order to evade an authorship attributor it may take a number of transformations and so the evaded document is liable to look more different from the original compared to the obfuscated version with a few modifications. \padmini{isn't this only true for mutantx? this thinking is just not clear throughout the paper. for the other 2 there is only one version and it evades or does not. with mutant x??? \asad{So the point is that the other two methods don't make a fixed number of transformations. It depends on the input text. Some text align with more rules and have more transformations than others (for two PAN methods)}}
%
%Thus we also expect evaded documents to have less smooth text than obfuscated documents.
%
%This in turn implies that we expect detection of obfuscated documents to be relatively more difficult.
%
%Moreover, in the real world we would expect users who value their privacy and seek obfuscation to publish only if the obfuscated document has evaded known attribution classifiers.
%
It is noteworthy that evaded documents are the realistic goal from a privacy seeking user's perspective.
Thus, in addition to assessing detection for obfuscated data, we also assess detection for evaded data for both (\amt and \blogs) obfuscated datasets.

\begin{table}
\scriptsize
\centering
\begin{tabular}{|c|c|c|c|c|}
\hline
\textbf{Dataset}                          & \multicolumn{2}{c|}{\textbf{Train}} & \multicolumn{2}{c|}{\textbf{Test}} \\ \cline{2-5}
                                 & \textbf{Original}    & \makecell{\textbf{Obfuscated}/ \\[-0.5ex]\textbf{Evaded}}   & \textbf{Original}   & \makecell{\textbf{Obfuscated}/ \\[-0.5ex]\textbf{Evaded}}   \\ \specialrule{1.5pt}{1pt}{1pt}
\makecell{\amt \\[-0.5ex]obfuscated}   & 431         & 80           & 268        & 47           \\ \hline
\makecell{\amt \\[-0.5ex]evaded}      & 236         & 29           & 235        & 30           \\ \hline
\makecell{\blogs \\[-0.5ex]obfuscated} & 2000        & 292          & 1900       & 277          \\ \hline
\makecell{\blogs \\[-0.5ex]evaded}     & 700         & 71           & 1000        & 113           \\ \hline
\end{tabular}
\caption{Number of original and obfuscated/evaded documents in train and test sets of each of the four datasets}\label{dataset_stats}

\end{table}

% \end{comment}
% \presubsub
\subsubsection{Final datasets and cross validation}

For each dataset, we use the original and obfuscated/evaded documents by 5 (out of 10) authors for training and the documents by the remaining 5 authors for testing.
%
% We split our dataset in this manner to get rid of any possible original-obfuscated document mapping on the basis of author.
%
An important decision to make here is with regards to the distribution of original documents versus obfuscated (or evaded) ones in our training and testing set. 
We have no real world scenario to mimic in that we have not encountered any real world use of automated obfuscators and their outputs.
Therefore we make the datasets under a reasonable assumption that original documents are in the vast majority.
Thus we create train and test datasets with roughly 85\% - 90\% original documents.
We do this by adding more original documents from each data collection to achieve these proportions.  
Overall, we ensure that 
each author's documents (of any version) are all either in the train or in the test set.
%
% (see table \ref{dataset_stats} for exact numbers).
%
Table \ref{dataset_stats} provides the size details for our obfuscated and evaded datasets.
%Finally we have the following four datasets. \textit{ \amt obfuscated} containing 699 original and 127 obfuscated documents, \textit{ \amt evaded} containing 471 original and 59 evaded documents, \textit{ \blogs obfuscated} containing 3900 original and 569 obfuscated documents and \textit{ \blogs evaded} containing 1700 original and 184 evaded documents.
%
%Table \ref{dataset_stats} shows the exact number of documents in train and test set for each of these four datasets.

% \presub
\subsection{Obfuscation Detection Methods}
\postsub

\subsubsection{Methods from our architecture}\label{mfoa}

We propose a total of 160 distinct architectures (see \ref{all_architectures}). %spanning over  three dimensions of Figure \asad{xx}.
Testing these for each of the four datasets, we conduct a total of 640 distinct obfuscation detection experiments.
In addition, we explore three other methods; the first is inspired by recent research to detect synthetic text.
The other two were used previously to detect manually obfuscated documents.

\subsubsection{Other Methods}\label{other_methods}

\vspace{0.05in} \noindent \textbf{1) GLTR \cite{gehrmann2019gltr}}.
The authors present a tool to help humans distinguish between original and machine generated synthetic text.
Their tool uses pretrained language models to extract word likelihoods and presents their plot to humans making the decision.
Their best model uses \gptsmall language model to extract word ranks which are then put in 4 unequal range bins: 0-10, 10-100, 100-1000 and greater than 1000.
We test the same method on our obfuscation detection problem. % where the key difference is that the modified text has components of its original human generated text.
%This method is inspired by recent research on the related yet distinct problem of differentiating between human and fully synthetic text generated using language models.
%
%
%Note that this method is a special type of architecture presented in \ref{all_architectures}.
%
For obfuscation detection, we tested this feature set with linear \svm and \rfc.
Among these, linear \svm performed the best.
%
% For evaluation, we used the code made available by the authors. 

\vspace{0.05in} \noindent \textbf{2) Character trigrams + \knn \cite{juola2012detecting}}.
This paper presents a stylometric approach to detect \textit{manually} obfuscated documents.
Firstly, they preprocess each document by unifying case and separating punctuation (e.g., !! becomes ! !).
Then they represent each document using character tri-grams.
Finally the classification is performed using \knn with normalized cosine distance.
We implement this approach to evaluate it against automated authorship obfuscaters.

\vspace{0.05in} \noindent \textbf{3) Writeprints + \svm \cite{afroz2012detecting}}.
This paper also uses stylometric features and is also focused on detecting manual obfuscation.
It is also similar to the approach in \cite{shahid2017accurate}, which uses stylometric features to perform spun document detection.
Afroz et al. tested with 3 different feature sets of which writeprints was the best.
Their writeprints feature set contains character related features (e.g., total characters, percentage of uppercase letters etc.), word related features (e.g., total words, frequency of large words etc.) and other features including frequencies of digits, special characters, function words etc.
They use this feature set with \svm (using \textit{poly} kernel) to perform obfuscation detection.
On our datasets, we found linear \svm to be working better than the polynomial one, so we report results with linear \svm.
We implement this writeprints approach with linear \svm as our final detector.

%\input{Sections/6_tables}

% \begin{figure*}[htpb]
%     \centering
%     \vspace{-20pt}
%     \subfloat[width=0.1\textwidth, trim={1.7cm 1.7cm 1.7cm 1.7cm}, clip][a]{\includegraphics{images/Classifier_evaded.pdf}\label{<figure1>}}
%     \subfloat[width=0.33\textwidth, trim={1.7cm 1.7cm 1.7cm 1.7cm}, clip][b]{\includegraphics{images/Classifier_evaded.pdf}\label{<figure2>}}
%     \vspace{-15pt}
%     \caption{.}
%     \label{figure:comunities}
% \end{figure*}

% \begin{figure}
%   \centering
%   \includegraphics[width=\linewidth]{images/LanguageModels_evaded.pdf}
%   \caption{Notched box plots of F1 scores for each language model for all experiment combinations (40) across the 2 evaded datasets.}\label{lm}
% \end{figure}

% \presec
\section{Results}
\label{sec: results}
\postsec
\textbf{Summary trends:} After averaging we find that for obfuscation detection, 25\% of all 160 architectures achieve F1 score greater than 0.76, 50\% achieve F1 score greater than 0.72 and a high 75\% of them were able to achieve F1 score greater than 0.52.
% Across all 640 experiments (160 parameter combinations for each dataset), we find that 25\% ?? 50\% achieve an F1 score greater than 0.65
% while 75\% achieve an F1 score of 0.5 or more. has to be dataset independent.  recalculate.
%
% Thus a higher percentage of our proposed architectures were able to achieve F1 score greater than 0.5.

\begin{figure}[!b]
  \centering
  \includegraphics[width=\linewidth]{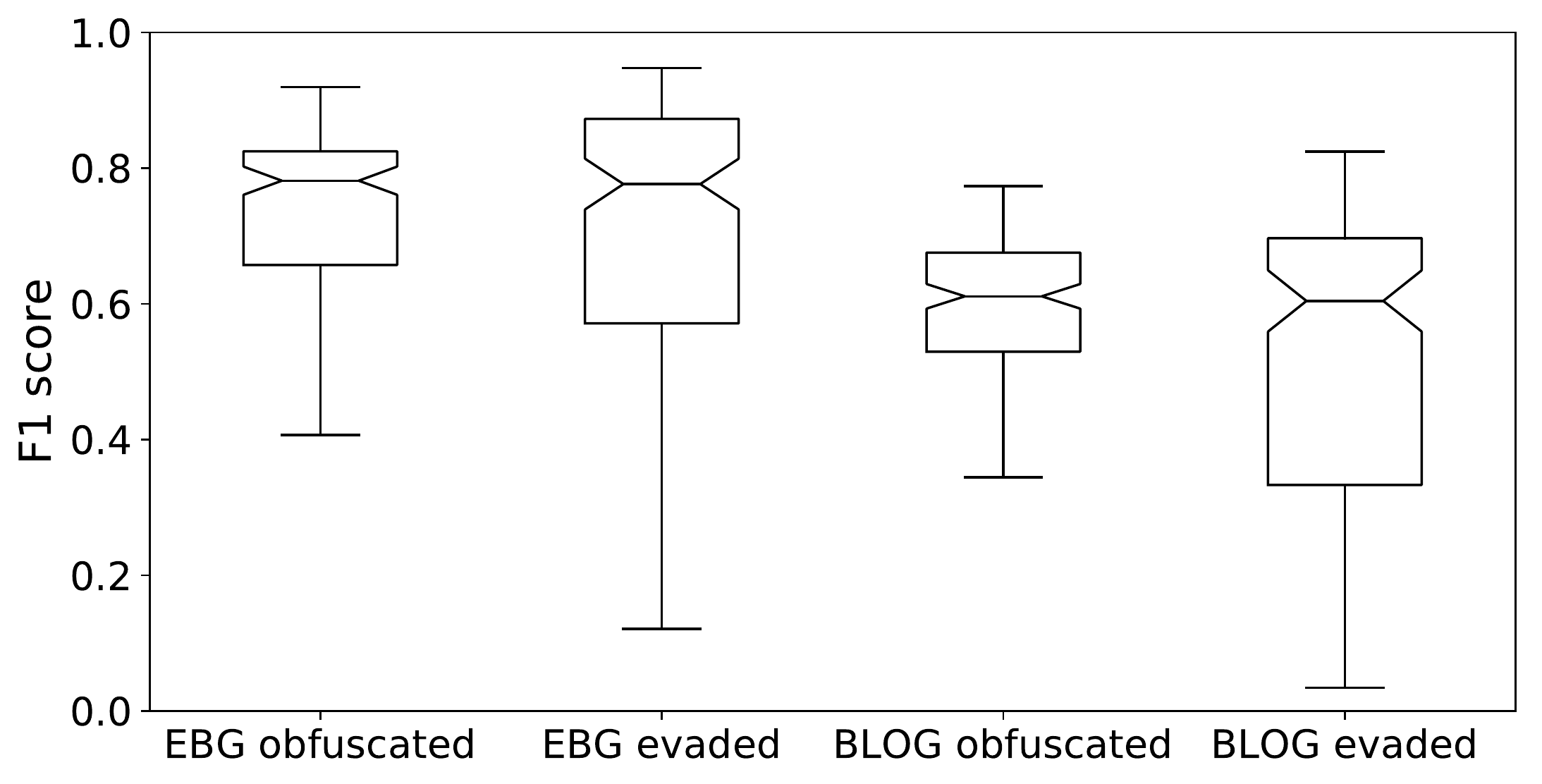}
  \caption{Notched box plots for obfuscation detection F1 scores  using all 160 architectures for each dataset.}\label{datasetwise}
\end{figure}

Figure \ref{datasetwise} summarizes the performances of all 160 different architectures across the four datasets.
We see that obfuscation detection is easier in \amt than in \blogs with median \amt F1 scores being significantly better than median \blogs F1 scores (notches do not overlap \cite{krzywinski2014points}).
This can be explained by the fact that \amt contains scholarly articles that are relatively more consistent in their smoothness than blogs. 
%
% \padmini{do you mean by consistency more smooth?? unclear.}\asad{since these are scholarly documents, they have higher smoothness across the whole datasets -> consistently smooth across the corpus}
%
This likely makes it easier to pick up on the difference in smoothness caused by obfuscated documents in \amt than in \blogs.
% differentiate between original and obfuscated documents on the basis of smoothness .
%
We can also see that evaded documents achieve a higher maximum F1 score than obfuscated documents.
This confirms our intuition presented in \ref{obfandevaded}, that evaded documents are likely to be less smooth and therefore easier to detect than obfuscated documents.
However, we also see that F1 scores for evaded datasets are less stable (greater box size) than obfuscated datasets.
We believe that this is due to the fact that there are fewer documents in evaded datasets as compared to their respective obfuscated datasets (see Table \ref{dataset_stats}).

% This plot also confirms the inference that obfuscation detection on \amt dataset in relatively easier than in \blogs dataset.
% %
% In fact the median F1 scores for EBG datasets are significantly better than median F1 scores for \blogs datasets (notches do not overlap \cite{krzywinski2014points}).
% %
% This figure also shows that evaded datasets are able to achieve higher F1 scores than obfuscated datasets but F1 scores for evaded datasets vary more (greater box size) than obfuscated datasets.

\noindent \textbf{Performance evaluation: }In terms of architecture selection, instead of choosing randomly across 160 architectures, we make the following assumption.
We assume that the adversary is knowledgeable about the various choices, tests these alternatives and employs the best configuration.
% Instead of a random selection for architecture we make a strong assumption that the adversary is knowledgeable about the various choices, tests these alternatives and employs the best configuration. 
%This indicates that ...
%which shows that there are relatively less number of experiments having F1 score less than 0.5 as compared to experiments having f1 score greater than 0.5.
Thus, we present results for the best models, based on F1 scores for obfuscation detection, achievable by the adversary (Table \ref{main_results}).
%Adversary trying to perform obfuscation detection has access to 4 language models each of which give 2 types of outputs.
%
%These outputs can be used to extract 4 different features sets which can be used with any of 5 different classifiers.
%
%So in total, adversary has access to 160 unique architectures and can pick the best one.
%
%Hence from our architectures, we show results only for the top performing ones.
%
% The best architectures are chosen on the basis of F1 score.
%
Table \ref{main_results} also presents results for the three additional methods presented in section \ref{other_methods}.
Our best BERT and GPT2 combinations outperform all other methods across each of the four datasets in F1 score and recall.
Along with (GLTR + SVM) these achieve the best precision for the \amt datasets.
In \blogs obfuscated, GLTR based method achieves the highest precision whereas in \blogs evaded both the GLTR based method and character trigrams method top the chart - however in each case with a sizeable penalty paid in recall and therefore in F1 score.
In summary, we see that using the best of methods the adversary can detect evaded and obfuscated documents with F1 score of 0.77 or higher (average 0.87 across datasets) which indicates that the tested state-of-the-art obfuscators are far from stealthy.

% Table \ref{main_results} also shows that obfuscation detection is easier in \amt dataset than in \blogs dataset.
% %
% This can be explained by the nature of documents in each dataset. %
% \amt dataset comprises of scholarly articles which are relatively more consistent in smoothness than blogs.
% %
% Hence, it is easier to differentiate between original and obfucated documents on the basis of smoothness in \amt dataset than in \blogs dataset.
% %
% We can also see that evaded documents achieve higher F1 scores than obfuscated documents.
% %
% This confirms our intuition presented in \ref{obfandevaded}, that evaded documents are less smooth and therefore easier to detect than obfuscated documents.

%Hence, if it is important not to misclassify original documents as obfuscated (false positives), then choosing the model with high precision is right.
%
%On the other hand, if you need to successfully detect obfuscated documents without caring about the misclassification of original documents  i.e., less false negatives, then using the model with high recall makes more sense.

\begin{table}[!t]
\scriptsize
\begin{tabular}{|c|p{3cm}|c|c|c|}
\hline
\textbf{Dataset}                         & \multicolumn{1}{c|}{\textbf{Models}}             & \textbf{P} & \textbf{R} & \textbf{F1}   \\ \specialrule{1.5pt}{1pt}{1pt}
\multirow{5}{*}{\makecell{\amt \\[0ex]obfuscated}}& \bertlarge + ranks + VGG-19 + \rfc      & 1.00      & 0.85   & \textbf{0.92} \\ \cline{2-5} 
                                & \bertlarge + ranks + VGG-19 + \svm      & 0.98      & 0.83   & \textbf{0.90} \\ \cline{2-5}
                                & GLTR + \svm                              & 1.00      & 0.70   & 0.83 \\ \cline{2-5} 
                                & Writeprints + \svm           & 0.67      & 0.38   & 0.49 \\ \cline{2-5} 
                                & Character trigrams + \knn     & 0.64      & 0.15   & 0.24 \\ \specialrule{1.25pt}{1pt}{1pt}
\multirow{5}{*}{\makecell{\amt \\[0ex]evaded}}    & \bertlarge + probs + bins(0.010) + \nueralnetworks   & 1.00      & 0.90   & \textbf{0.95}  \\ \cline{2-5} 
                                & \bertsmall + probs + VGG-19 + \naivebayes  & 1.00      & 0.90   & \textbf{0.95} \\ \cline{2-5} 
                                & GLTR + \svm                              & 1.00      & 0.80   & 0.89 \\ \cline{2-5} 
                                & Writeprints + \svm           & 0.79      & 0.63   & 0.70 \\ \cline{2-5} 
                                & Character trigrams + \knn    & 1.00      & 0.17   & 0.29 \\ \specialrule{1.25pt}{1pt}{1pt}
\multirow{5}{*}{\makecell{\blogs \\[0ex]obfuscated}}& \bertsmall + probs + VGG-19 + \nueralnetworks                  & 0.85      & 0.71   & \textbf{0.77} \\ \cline{2-5} 
                                & \bertsmall + probs + VGG-19 + \svm             & 0.79      & 0.74   & \textbf{0.77} \\ \cline{2-5} 
                                & GLTR + \svm                                        & 0.92      & 0.40   & 0.56 \\ \cline{2-5} 
                                & Writeprints + \svm           & 0.71      & 0.41   & 0.52 \\ \cline{2-5} 
                                & Character trigrams + \knn    & 0.41      & 0.50   & 0.45 \\ \specialrule{1.25pt}{1pt}{1pt}
\multirow{5}{*}{\makecell{\blogs \\[0ex]evaded}}  & \gptlarge + ranks + VGG-19 + \naivebayes      & 0.82      & 0.83   & \textbf{0.83} \\ \cline{2-5} 
                                & \bertsmall + probs + VGG-19 + \nueralnetworks             & 0.79      & 0.81   & \textbf{0.80} \\ \cline{2-5} 
                                & GLTR + \svm                                        & 0.86      & 0.55   & 0.67 \\ \cline{2-5} 
                                & Writeprints + \svm           & 0.84      & 0.62   & 0.71 \\ \cline{2-5} 
                                & Character trigrams + \knn    & 0.86      & 0.50   & 0.63 \\ \specialrule{1.25pt}{1pt}{1pt}
\end{tabular}
\centering
	\caption{Obfuscation detection results (P: precision, R: recall, F1: F1 score).}\label{main_results}
	
\end{table}

\begin{figure*}
  \centering
  \includegraphics[width=\linewidth]{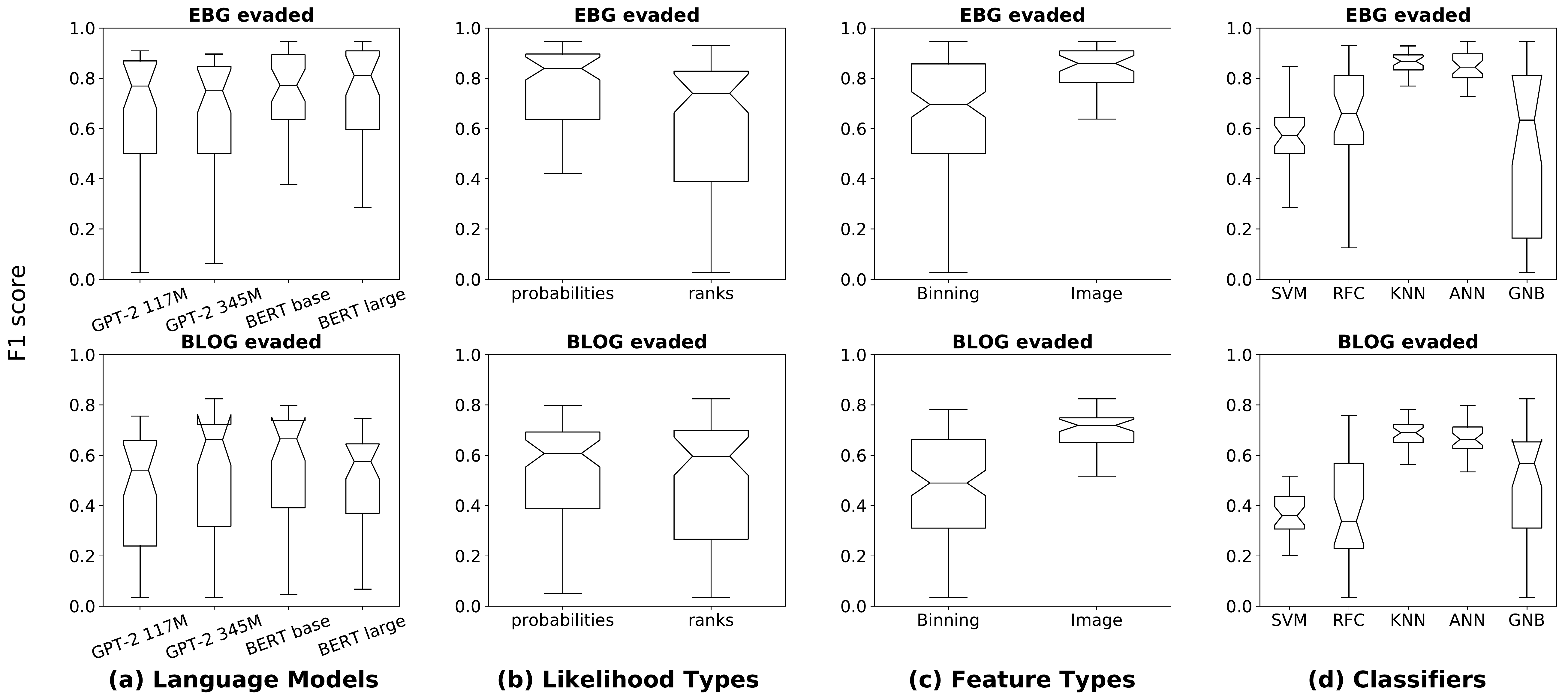}
  \caption{Notched box plots of F1 scores for all dimensions across the two evaded datasets. For each dataset every notched boxplot in (a) is generated from 40 experiments (experiments correspond to architectures), (b) is generated from 80 experiments, (c) is generated from 120 experiments for binning and 40 for image whereas (d) is generated from 32 different experimental combinations. }\label{all_dimentions}
  
\end{figure*}

\subsection{Detector Architecture Choices Analysis}

Now we analyze the effect of different choices made within each of the three dimensions depicted in Figure \ref{pipeline}.
As mentioned earlier, for a privacy seeking user evading author attribution is more important than just obfuscation.  %s will only publish evaded documents rather than just somewhat obfuscated documents. people trying to obfuscate documents would only publish if it evades some attribution system.
%
%Hence, in real world we expect to encounter evaded documents rather than obfuscated documents.
%
So, in this section we present architecture analysis results only for evaded datasets involving 320 experiments (160 each for \amt evaded and \blogs evaded).
% \presubsub
\subsubsection{Dimension 1: Language model \& output type}
\postsubsub
%In our experiments, we tested with 4 different langauge models i.e., bert base, bert large, GPT-2 117M and GPT-2 345M.
%
Figure \ref{all_dimentions} (a) presents notched box plots comparing distribution of F1 scores achieved by language models across both datasets.
%by experiments using these langauge models .
%
In \amt evaded, BERT language models achieve higher maximum F1 score (0.95) than GPT-2 (0.90 - 0.91).
On the other hand, in \blogs evaded, \gptlarge achieves higher maximum F1 score (0.83) than others (0.75 - 0.80).
% \bertlarge achieves the highest F1 score of 0.91 and 0.93 on \amt obfuscated and evaded datasets respectively.
% %
% The highest F1 score for \blogs obfuscated is achieved by \bertsmall (0.76) and for \blogs evaded by \gptlarge (0.81).
%
Relatively, BERT shows greater consistency in F1 score (box size) than GPT-2 in both datasets.
We believe that the bidirectional nature of BERT helps in capturing context and consequently smoothness better than GPT-2 which is uni-directional.
%In terms of F1 score variability (box size) we see that for \amt evaded, BERT shows greater consistency than when using GPT-2.
%
%In contrast, all models are similar in box size for \blogs datasets.
%
% The only significant difference found is where \amt obfuscated, where median F1 for \bertlarge is significantly better than for \gptsmall.
%
%\vspace{0.05in} \noindent \textbf{2) Ranks versus probabilities as output:}
%We tested with two output types i.e., probabilities and ranks
%

While the difference in maximum F1 score between ranks and probabilities is slight for each dataset (Figure \ref{all_dimentions} (b)) 
%shows that in \amt evaded, output probabilities achieve higher maximum F1 score (0.95) than output ranks (0.93) whereas in \blogs evaded ranks achieve higher maximum F1 score (0.83) than probabilities (0.80).)
%
% With obfuscated datasets using ranks is the best (0.91) for \amt whereas probabilities win for \blogs (0.76).
%experiment,  were able to achieve the highest F1 score of 0.76.
%
box sizes show the spread in F1 scores is smaller with probabilities than with ranks.
Upon further investigation, we find that experiments which use probabilities with image based features have an inter-quartile range of 0.05 and 0.1 for \amt and \blogs respectively whereas for experiments using probabilities with binning based features, this range is 0.32 for both datasets.
On the other hand, inter-quartile range for experiments using ranks with image based features is 0.08 and 0.05 for \amt and \blogs whereas for experiments using ranks with binning based features, this range is 0.49 and 0.42 respectively.
This shows that for both datasets, greater variation in F1 scores for ranks as compared to probabilities is caused by binning based features.
We believe that binning ranks with fixed bin sizes (10, 50, 100) is less stable for both BERT and GPT-2 which have different limits of ranks - this could account for the larger inter-quartile range using ranks.

% \presubsub
\subsubsection{Dimension 2: Feature type}
\postsubsub
% We tested with two different feature representations i.e., binning and VGG-19 features.
%
%Notched box plots in figure \ref{ft} show the comparison of F1 scores between binning and image based feature representations across all four datasets.
%
The box sizes in Figure \ref{all_dimentions} (c) show that image based features exhibit strikingly greater stability in F1 scores than binning based features.
Image based features also achieve significantly higher median F1 score than with binning for both datasets.
This can in part be explained by the observation stated earlier that some bin size choices tested perform much worse than others because of not being fine-tuned.
There is no difference between feature types in maximum F1 score for \amt whereas in \blogs, image based feature achieve somewhat higher maximum F1 score (0.83) than binning based features (0.78). 
We believe that the reason why image based features work so well is that VGG-19, the image model we use to extract features, is powerful enough to recognize the slopes in plots which represent the smoothness in our case.
% for \amt obfuscated, binning based features is the best (highest F1 score of 0.91) whereas for \blogs obfuscated, image based features is the best (highest F1  of 0.76).
%
% Across both datasets, the box sizes show that experiments using image based feature representation achieve less variable F1 scores than binning based features which can be explained with the different choices of bin sizes.
%

% \begin{figure}
%   \centering
%   \includegraphics[width=\linewidth]{images/Feature_type_evaded.pdf}
%   \caption{Notched box plots of F1 scores for feature types for all experiment combinations (binning based: 120, image based: 40) across the 2 evaded datasets.}\label{ft}
% \end{figure}

% \begin{figure}
%   \centering
%   \includegraphics[width=\linewidth]{images/Classifier_evaded.pdf}
%   \caption{Notched box plots of F1 scores for classifiers used for all experiment combinations (32) across the 2 evaded datasets.}\label{c}
% \end{figure}
% \presubsub
\subsubsection{Dimension 3: Classifier}
\postsubsub
% For classification, we used a range of different classifiers including GNB, KNN, ANN, SVM, RFC and DC(mahalanobis).
%
% Figure \ref{c} shows the range of F1 scores achieved by each classifier across both datasets.
%
% For \amt obfuscated, \knn and \nueralnetworks achieve the highest F1 score of 0.91 whereas for \amt evaded, \nueralnetworks and \rfc achieve the highest F1 score (0.93).
% %
% For \blogs obfuscated, \svm achieves the highest F1 score of 0.76 and for \blogs evaded, \knn and \naivebayes achieve the highest F1 score of 0.81.
%
%
Figure \ref{all_dimentions} (d), shows that for \amt, \nueralnetworks and \naivebayes achieve higher maximum F1 score (0.95), whereas for \blogs, \naivebayes achieve higher maximum F1 score (0.83).
\knn and \nueralnetworks consistently achieve far more stable F1 scores than other classification methods.
%
% Moreover, we see that in \blogs, 1st quartile of \knn and \nueralnetworks is higher than the 3rd quartile of other classification methods.
%
In both datasets, \knn achieves significantly higher median F1 score than other classification methods.
\nueralnetworks also follows the same pattern with the exception of \naivebayes in \blogs evaded.
We believe that the reason why \knn and \nueralnetworks achieve relatively high and stable performance is in their nature of being able to adapt to diverse and complex feature spaces.

\presub
\subsection{Takeaway}
\postsub
In summary we conclude that BERT with probabilities is a good choice for dimension 1. (We remind the reader that in contrast, in the area of synthetic text detection \cite{gehrmann2019gltr} GPT-2 had the edge over BERT).
Image based features are a clear winner in dimension 2 while \knn and \nueralnetworks are the best candidates for dimension 3.
Key to note as well is that the top performing architectures in Table \ref{main_results} differ across datasets indicating the need for dataset specific choices.
%
%This indicates that performance of an architecture is dependent on the type of documents present in the dataset. 
%
%Hence, any adversary trying to detect obfuscation should preferably train all these architectures and then pick the best one.

% \presub
\subsection{Insights}
\begin{figure}
  \centering
  \includegraphics[width=\linewidth]{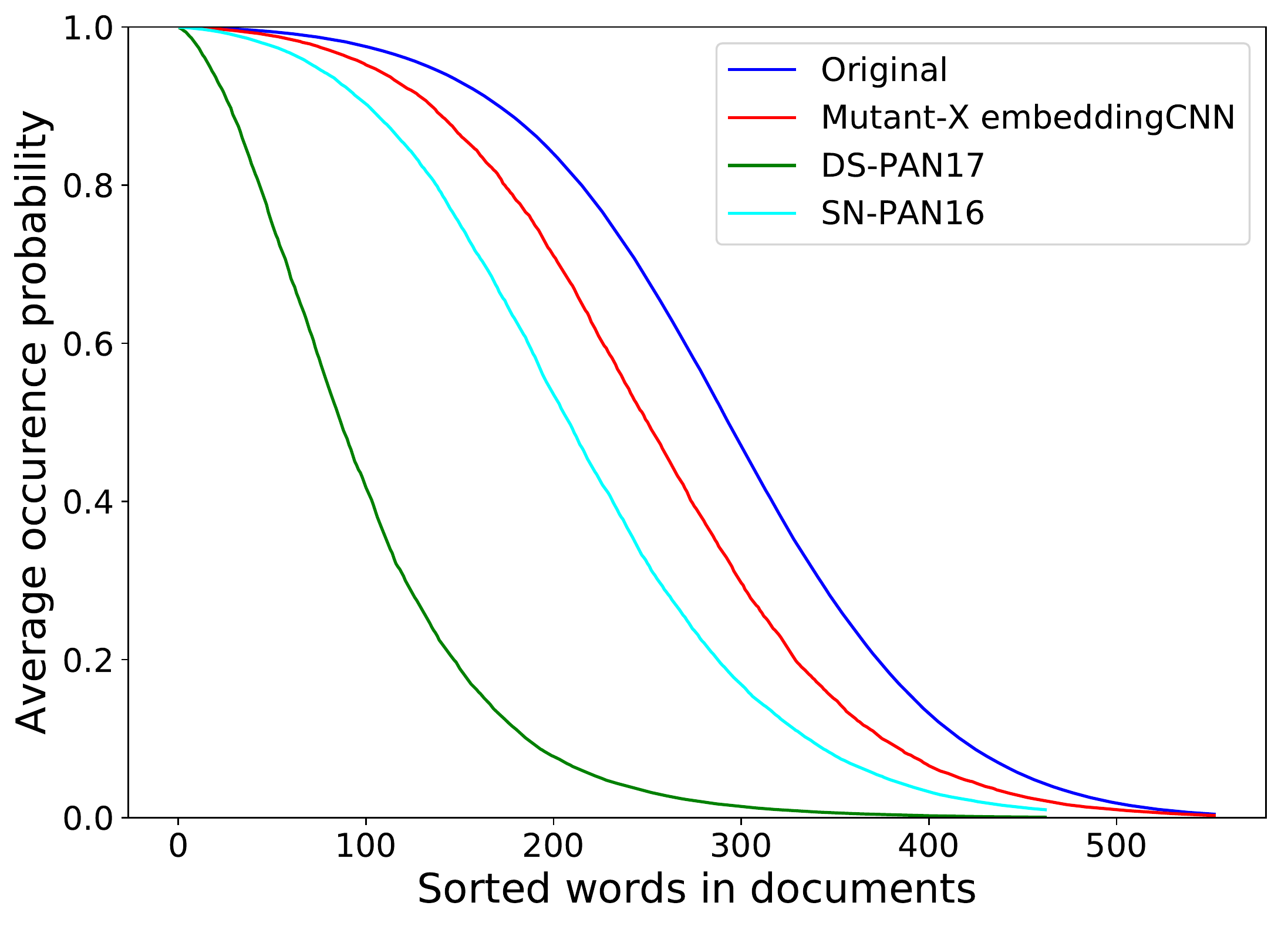}
  \caption{Comparison between different obfuscators and original documents on the basis of average sorted probabilities extracted by \bertsmall for \amt obfuscated dataset.}\label{intuit}
\end{figure}

\postsub
Figure \ref{intuit} validates our intuition from Section \ref{sec: approach} that the text generated by obfuscators is less smooth than the original text. 
Using \amt obfuscated dataset and \bertsmall for illustration, we first sort words in a document by estimated probability and plot average probability at each rank. 
The steeper the fall in the curve, the lower the smoothness of text.
This plot shows that original documents are generally more smooth than obfuscated documents.
The average detection error rates (Mutant-X embeddingCNN: 0.72, SN-PAN16: 0.48, and DS-PAN17: 0.07) are also consistent with the plot. 
These results show that Mutant-X is the most stealthy obfuscator while DS-PAN17 is the least stealthy obfuscator.

% For this plot, we first extract the occurrence  of words in each document using \bertsmall and sort them for all the documents .
% %
% Then we average these sorted probabilities for different obfuscator generated and original documents and plot them.
% %
% The steeper the fall, the greater the average number of low probability words in documents and hence lower the smoothness of text.
% %
% This plot shows that original documents are more smooth on average whereas \dspan generates the least smooth documents.

% \presec
\section{Conclusion}
\postsec
In this paper, we showed that the state-of-the-art authorship obfuscation methods are not stealthy.
We showed that the degradation in text smoothness caused by authorship obfuscators allow a detector to distinguish between obfuscated documents and original documents.
Our proposed obfuscation detectors were effective at classifying obfuscated and evaded documents (F1 score as high as 0.92 and 0.95, respectively).
Our findings point to future research opportunities to build stealthy authorship obfuscation methods. 
We suggest that obfuscation methods should strive to preserve text smoothness in addition to semantics. 
%%One limitation is that we have not studied characteristics influencing stealthiness of individual obfuscators. This is also left for future research.

% text smoothness in individual documents there may be other signatures left by obfuscators such as ones that might be observed only in aggregate from groups of documents.
%that make obfuscation detectable such as patterns across groups of documents. 
%Since our detection architectures use language models to assess smoothness, if an obfuscator was developed which incorporates the smoothness consistency feedback using the same language models, performance of our architectures would probably decrease.
% \padmini{add 1 sentence limitations}\asad{done}

%
%We think that obfuscation detection is potentially a more challenging problem than synthetic text detection because the characteristics of obfuscated text depend on both the original style of text and obfuscation system used whereas in synthetic text, it depends on just the language model.
%
%We use 160 unique architectures along with some other methods for evaluation.
%

\bibliography{Asad,mutantx}
\bibliographystyle{acl_natbib}

\end{document}